%% file: iclr2027_conference.tex
\documentclass{article} 
\usepackage{iclr2027_conference,times}

\input{math_commands.tex}

\usepackage{hyperref}
\usepackage{url}
\usepackage{graphicx}
\usepackage{float}

\title{Learning Administrative Boundary-Aware Location Encoders via Reverse Geocoding}

\author{Lakshay Sharma  \\
Instacart \\
New York, NY, USA \\
\texttt{lakshay.sharma@cims.nyu.edu} \\
\And
Alex Marin \\
Thomson Reuters Labs \\
Seattle, WA, USA \\
\texttt{amarin@uw.edu}
}

\iclrfinalcopy 
\begin{document}

\maketitle

\begin{abstract}
    Location encoders learn compact representations of latitude--longitude coordinates for downstream geospatial prediction tasks. Existing approaches are pretrained using continuous or naturally occurring supervision, such as imagery, climate variables, or species distributions, and therefore do not explicitly capture administrative boundaries or place names, despite their strong influence on many real-world geospatial phenomena. We introduce, to our knowledge, the first method for learning administrative boundary-aware location encoders, using reverse-geocoded text as a self-supervised training signal. Our approach can be applied to existing location encoder architectures and consistently improves their downstream performance. We further introduce RevGeoRQ, a custom architecture tailored to reverse-geocoding supervision, which yields additional gains and exceeds contemporary location encoders on several downstream tasks. Finally, the use of human-interpretable reverse-geocoded strings and residual-quantized semantic IDs provides a more interpretable representation of geographic location than existing approaches. Code and model weights to be released soon.
\end{abstract}

\section{Introduction}

Location encoders have emerged as a foundational building block for modern data-driven geographic information systems, enabling models to convert raw latitude–longitude coordinates into rich, high-dimensional representations. Learning effective such spatial representations is critical across a wide spectrum of downstream tasks, ranging from predicting species distributions and environmental variables to optimizing urban planning, route navigation, and localized economic forecasting. By replacing hand-engineered spatial features with expressive pre-trained neural embeddings, location encoders allow predictive models to better reason about distance, regional continuity, and geographic context directly from raw coordinate inputs, reducing the need for time-consuming implementation.

Despite their success, training general-purpose location encoders that transfer effectively across diverse downstream applications remains challenging. Existing representation learning strategies typically rely on continuous pretraining targets, such as multispectral satellite imagery, climate grids, or naturally occurring species presence logs. While these signals capture smooth environmental gradients and various visual patterns, they largely overlook the discrete, structured boundaries that define human geography. However, real-world geospatial phenomena are heavily shaped by administrative borders, regional nomenclature, and socio-political divisions - for example, boundaries between urban and natural terrain are often artificial, informed by administrative boundaries - exactly the type of information that purely physical or visual pretraining signals fail to encode explicitly. As a result, current encoders often struggle on downstream tasks where administrative context and localized place identity play a decisive role.

To bridge this gap, we use reverse-geocoded text as a rich, scalable, and relatively cheap signal for developing self-supervised location encoders. Reverse geocoding translates raw spatial coordinates into structured text strings containing hierarchical place names, administrative units, and other localized metadata. By training encoders to predict or align with these textual representations in a self-supervised manner, models can internalize both continuous spatial topology and discrete human-centric boundaries without requiring costly manual annotations. This supervision source is applicable across diverse encoder architectures, providing a fine-grained, semantically rich representation of geographic space that aligns well with human mental models of geography. The novel self-supervised training is one of the main contributions of our work, alongside a novel architecture tailored to reverse-geocoding supervision. As an additional contribution, we show evidence that reverse geocoded output improves interpretability in downstream systems.

The rest of the paper is structured as follows. We discuss past work and how our contributions complement these past approaches in section \ref{sec:related-work}. Section \ref{sec:methodology} contains the description of our novel model architecture and the location encoder usage. Our main experimental results on downstream tasks are discussed in section \ref{sec:experimental_setup}, with a detailed ablation study discussed in section \ref{sec:ablations}. We perform further analysis of the interpretability of our embeddings in section \ref{sec:embedding-analysis}. Finally, we briefly discuss limitations to our approach in section \ref{sec:limitations}, and conclude with future work in section \ref{sec:conclusions}.

\section{Related Work}
\label{sec:related-work}

Location encoders project continuous geographic coordinates (latitude and longitude) into high-dimensional representations for downstream prediction tasks. Early approaches relied on non-parametric sinusoidal encodings or explicit spherical coordinate parameterizations, such as Sphere2Vec \cite{mai2023sphere2vec}, to preserve great-circle distances without planar distortion. Modern representation learning paradigms instead pretrain location encoders in a self-supervised manner by grounding coordinate representations in co-located modalities. Multimodal frameworks such as GeoCLIP \cite{vivance2023geoclip} and SatCLIP \cite{klemmer2025satclip} contrastively align coordinate embeddings with remote-sensing or street-level imagery, while ecological models like BioCLIP \cite{stevens2024bioclip} use hierarchical taxonomic occurrences. Further advancing image-coordinate alignment, RANGE \cite{dhakal2025range} enhances visual feature density through retrieval-augmented neural fields over imagery. However, because these methods rely exclusively on visual, physical, or ecological co-occurrence signals, they treat geographic space purely as a continuous physical manifold, failing to explicitly capture discrete, human-defined administrative structures and localized place names.

Integrating natural language with spatial data has gained momentum as Large Language Models (LLMs) demonstrate strong capabilities in geographic reasoning and spatial representation. Linear probing and cognition studies, such as \cite{gurnee2024space}, \cite{godey2024scaling}, and \cite{xia2026can}, show that LLMs can internalize structured global spatial layouts purely from textual metadata and local descriptions of points of interest. To take advantage of this latent spatial knowledge for downstream predictions, frameworks like GeoLLM \cite{manvi2024geollm} and LLMGeovec \cite{he2024geolocation} augment spatial coordinate prompts with auxiliary textual descriptions from OpenStreetMap or POI categories. While effective at unlocking spatial knowledge, these paradigms rely on heavy LLM backbones or expensive prompt-based inference at prediction time. In contrast, our approach uses reverse-geocoded text purely as a self-supervised pretraining signal, distilling rich administrative and textual spatial context directly into a compact, standalone location encoder without incurring LLM overhead at inference time.

Discrete quantization methods such as VQ-VAE \citep{van2017neural} and VQGAN \citep{esser2021taming} represent continuous high-dimensional spaces using learned discrete codebooks.This idea was later adapted successfully to generative retrieval and recommendation for retail \citep{rajput2023recommender}, video \citep{li2026tokens}, audio \citep{penha2025semantic}, and points of interest \citep{wang2025gnprsid,lv2026ros}. 

Reverse geocoding maps raw spatial coordinates to structured human-readable text, such as street addresses, municipal divisions, and administrative hierarchies. While commercial APIs, such as the Google Maps Geocoding API \cite{googlemapsgeocoding}, provide high-precision global reverse geocoding, their cost structure becomes prohibitively expensive when querying millions of coordinates required for large-scale machine learning pretraining. Conversely, free or open-source geocoding alternatives often suffer from spatial noise, incomplete boundary coverage, and positional inaccuracy, as systematically evaluated by Kounadi et al. \cite{Kounadi01032013}. To mitigate errors in raw geocoding outputs, probabilistic frameworks such as Yin et al. \cite{yin2023probabilistic} model candidate entity distributions over administrative zones to resolve geographic ambiguities. Rather than relying on online commercial APIs or static categorical lookups during downstream prediction, we use reverse-geocoded strings as a offline, self-supervised training signal. By learning to predict structured text representations during pretraining, location encoders internalize fine-grained administrative hierarchies directly from raw coordinates, distilling high-quality geographic context into a fast, standalone encoder without incurring runtime API expenses or latency. While we work primarily with free, off-the-shelf reverse geocoding systems, we expect that improving the quality of the reverse geocoder will only result in improvements in downstream performance, as shown in past work, such as \cite{kilic2023effects}.

\section{Methodology}
\label{sec:methodology}

\label{sec:methodology}

\subsection{Dataset construction}

Let \(c_i\) denote a geographic coordinate and \(R\) a reverse geocoder. Its reverse-geocoded string is

\begin{equation}
    s_i = R(c_i).
    \label{eq:reverse_geocoding}
\end{equation}

We format \(s_i\) as \texttt{city, admin1, admin2, country, continent}. For example, \((40.765685, -73.976239)\) maps to \texttt{Manhattan, New York, New York County, United States, North America}. To evaluate generalization across location encoder architectures, we apply our method to each model's native pretraining dataset. For model \(M_j\), coordinates set \(C_j=\{c_i\}\) are reverse geocoded to obtain \(S_j=\{s_i\}\), and the model is pretrained on the paired sets \((C_j,S_j)\). We use the \texttt{pycountry}\footnote{\url{https://pypi.org/project/pycountry/}}, \texttt{pycountry\_convert}\footnote{\url{https://pypi.org/project/pycountry-convert/}}, and \texttt{reverse\_geocoder}\footnote{\url{https://pypi.org/project/reverse-geocoder/}} libraries.

\subsection{Architecture}

\begin{figure}[h]
    \begin{center}
        \includegraphics[width=0.9\linewidth]{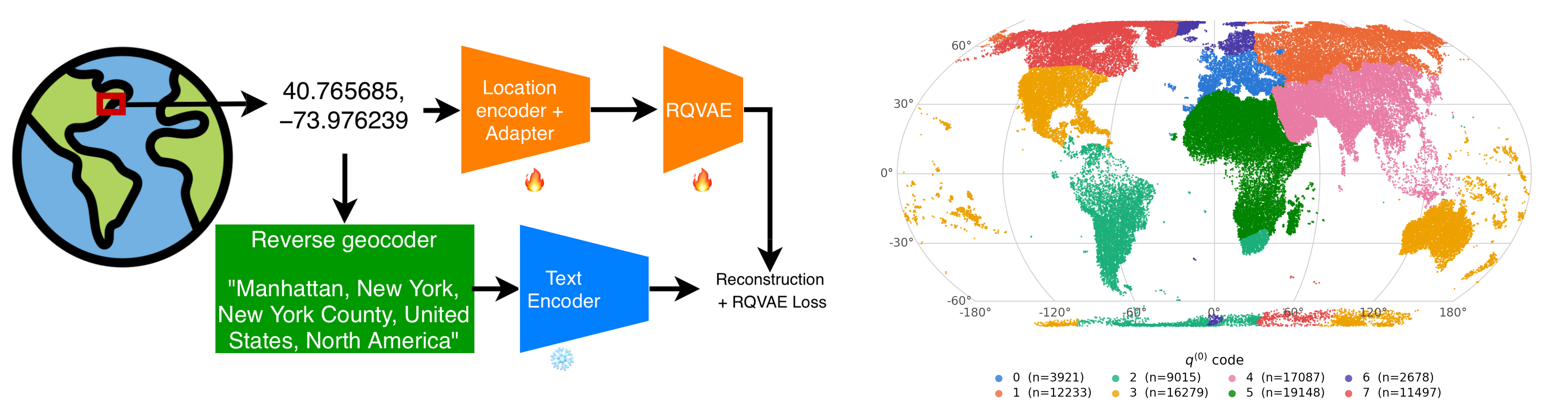}    
    \end{center}
    \caption{\textbf{Left:} Pretraining architecture. \textbf{Right:} Learned $q^{(0)}$ (level-0) codes from RQVAE Semantic IDs on S2-100K points}
    \label{fig:arch-and-q0}
\end{figure}

Our model architecture has three main components: a) a Location Encoder, b) a Text Encoder, and c) a Residual Quantized Variational Autoencoder (RQ-VAE), illustrated in Fig.~\ref{fig:arch-and-q0} (Left).

\subsubsection{Location encoder}

Given an encoder \(E\) and coordinates \(c_i=(\mathrm{lat}_i,\mathrm{lon}_i)\in\mathbb{R}^2\), the location embedding \(g_i\in\mathbb{R}^{d_g}\) is

\begin{equation}
g_i = E(c_i),
\qquad
E:\mathbb{R}^2\rightarrow\mathbb{R}^{d_g}.
\label{eq:location-embedding}
\end{equation}

The coordinates may be further transformed using spherical-harmonic, sinusoidal, or other frequency-based encodings \citep{klemmer2025satclip,russwurm2024geographic,dollinger2025climplicit}.

\subsubsection{Text encoder}

For a location with coordinates $c_i$, let $s_i$ denote its reverse-geocoded string, obtained according to Eq.~\ref{eq:reverse_geocoding}. The string $s_i$ is passed through a frozen pretrained text encoder $T$ to obtain the corresponding text embedding $t_i \in \mathbb{R}^{d_t}$:
\begin{equation}
    t_i = T(s_i),
    \qquad
    T : \mathcal{S} \rightarrow \mathbb{R}^{d_t},
    \label{eq:text-embedding}
\end{equation}
where $\mathcal{S}$ denotes the space of reverse-geocoded strings. We use the pretrained frozen Qwen3-Embedding (\cite{zhang2025qwen3}) models (0.6B and 4B variants) and use the unnormalized last-token hidden state from its final layer ($d_t = 1024 | 2560$, respectively).

\subsubsection{Residual-Quantized Variational Autoencoder}
\label{sec:rqvae}


Owing to the discrete nature of administrative boundaries introduced by reverse geocoding, and inspired by \cite{rajput2023recommender}, we use a Residual-Quantized Variational Autoencoder (RQ-VAE) \citep{zeghidour2021soundstream,lee2022autoregressive} to turn the location encoder output $g_i$ (Eq.~\ref{eq:location-embedding}) into a discrete, hierarchical code.
Unlike a standard RQ-VAE, which reconstructs its own input, we train this to use $g_i$ as input to construct the \emph{frozen} text embedding $t_i$ (Eq.~\ref{eq:text-embedding}) of the
location's reverse-geocoded string $s_i$ (Eq.~\ref{eq:reverse_geocoding}), as shown below.

A trainable linear projection $P:\mathbb{R}^{d_g}\!\to\!\mathbb{R}^{d_t}$ maps
location encoder backbones of differing dimensions into the text space, and an encoder produces the latent
$z_i=\mathcal{E}_q\!\big(P(g_i)\big)$. Residual quantization then runs over $m$
levels, each with its own codebook $\mathcal{C}_\ell=\{e_{\ell,k}\}_{k=1}^{K}$ of $K$
candidate codewords, where $k$ indexes the entries of $\mathcal{C}_\ell$. Let
$r_i^{(\ell)}$ denote the residual remaining after level $\ell$, starting from
$r_i^{(0)}=z_i$. Level $\ell$ finds the codebook entry nearest to the current
residual, records its index $q_i^{(\ell)}\in\{1,\dots,K\}$, and subtracts the
corresponding codeword $\hat e_i^{(\ell)}:=e_{\ell,q_i^{(\ell)}}$ to form the next
residual:
\begin{equation}
    q_i^{(\ell)}=\arg\min_{k\in\{1,\dots,K\}}\big\|r_i^{(\ell-1)}-e_{\ell,k}\big\|_2^2,
    \qquad
    r_i^{(\ell)}=r_i^{(\ell-1)}-\hat e_i^{(\ell)}.
    \label{eq:residual-quantization-reduced}
\end{equation}
Thus $q_i^{(\ell)}$ is the discrete code chosen at level $\ell$ and $\hat e_i^{(\ell)}$
the codeword it points to. Each level quantizes the residual left by the previous
ones, so the tuple of chosen codes $(q_i^{(1)},\dots,q_i^{(m)})$, which we call the
location's \emph{Semantic ID}, is a coarse-to-fine descriptor indexing up to $K^{m}$
distinct locations. The decoder reconstructs the text embedding from the summed
codewords, $\hat t_i=\mathcal{D}_q\!\big(\sum_{\ell=1}^{m}\hat e_i^{(\ell)}\big)$, and the
tokenizer is trained end-to-end with the loss
$\mathcal{L}=\mathcal{L}_{\mathrm{recon}}+\mathcal{L}_{\mathrm{rqvae}}$, where
$\mathcal{L}_{\mathrm{recon}}=\lVert t_i-\hat t_i\rVert_2^2$ and
$\mathcal{L}_{\mathrm{rqvae}}=\sum_{\ell=1}^{m}\big\lVert\operatorname{sg}[r_i^{(\ell-1)}]-\hat e_i^{(\ell)}\big\rVert_2^2+\beta\big\lVert r_i^{(\ell-1)}-\operatorname{sg}[\hat e_i^{(\ell)}]\big\rVert_2^2$.
Here $\operatorname{sg}[\cdot]$ denotes the stop-gradient operation
\citep{van2017neural} and $\beta$ weights the commitment loss; this objective jointly
trains the location encoder $E$, the projection $P$, the encoder $\mathcal{E}_q$, the decoder $\mathcal{D}_q$, and all $m$ codebooks. Specific details about the RQ-VAE design are provided in Appendix~\ref{app:rqvae}.

\section{Experimental Setup}
\label{sec:experimental_setup}

\subsection{Pretraining}

We experiment with five location encoder architectures: SatCLIP \citep{klemmer2025satclip}, GeoCLIP \citep{vivance2023geoclip}, SINR \citep{cole2023spatial}, Climplicit \citep{dollinger2025climplicit}, and a custom Reverse Geocoding Residual Quantized (RevGeoRQ) architecture.
\subsubsection{Adapting existing backbones}
\label{sec:pretrain-adapting-backbones}

Each model is pretrained on its native coordinate set, but we replace the original labels and training objective with our reverse-geocoded embeddings and objective. All parameters are randomly initialized to enable a direct comparison with native pretraining. During training, the location encoder and RQ-VAE outputs are aligned with the text encoder output using the loss described in Sec. \ref{sec:rqvae}. 

Since SatCLIP and GeoCLIP expose usable final-layer embeddings, the adapter is appended directly to them. In contrast, SINR and Climplicit have task-specific heads which are replaced by the adapter attached to the penultimate embedding layer. Our adapted SatCLIP model use uses \(L=10\) unlike the native backbone which uses \(L=40\) since we find that using \(L=40\) performs substantially worse in our setting; we investigate this in Sec. \ref{sec:ablations}.

To demonstrate gains with minimal architectural changes, we use the 0.6B Qwen encoder for backbone adaptation. Its smaller embedding dimension enables strong performance with only a lightweight adapter and no backbone modification. Larger Qwen3 variants are used in selected settings. Full pretraining details are provided in Appendix \ref{app:pretraining-details}, Table~\ref{tab:pretrain-params}.

\subsubsection{Custom architecture}
\label{sec:custom-architecture}

We introduce RevGeoRQ, a SatCLIP-based architecture optimized for our method, which uses Qwen3-Embedding-4B embeddings as reconstruction targets. A spherical-harmonic layer (\(L=10\)) maps coordinates to 100 dimensions, followed by a modified SirenNet \cite{russwurm2024geographic} with widths \(100 \rightarrow 1024 \rightarrow 2560 \rightarrow 2560\). Matching the 2560-dimensional Qwen output lets the representation feed directly into the RQ-VAE without an adapter. RevGeoRQ is pretrained on SatCLIP's S2-100K dataset using the settings in Sec. \ref{sec:pretrain-adapting-backbones}.

\subsection{Downstream fine-tuning}
\label{sec:downstream-finetuning}

We evaluate on four regression and three classification tasks. Classification includes country \citep{klemmer2025satclip}, ecoregion, and biome prediction (both from ~\cite{dinerstein2017ecoregion}); regression includes air temperature \citep{hooker2018global}, elevation and population density \citep{rolf2021generalizable}, and housing price prediction \citep{pace1997sparse}. We evaluate five variants on each downstream task:
\begin{enumerate}
    \item \textbf{Baseline}: reverse-geocoded text encoded by Qwen3-Embedding-0.6B and passed directly to the downstream MLP probe.
    \item \textbf{Native}: the original pretrained location encoder using author-released weights.
    
    \item \textbf{Text Residual Quantized (TextRQ)}: our adaptation of each backbone, using the \(\mathbb{R}^{1024}\) adapter output trained against Qwen3-Embedding-0.6B embeddings.
    
    \item \textbf{RevGeoRQ}: our custom architecture described in Sec. \ref{sec:custom-architecture}.

    \item \textbf{Combined}: concatenates \textit{Native} and \textit{TextRQ} embeddings for adapted backbones; for RevGeoRQ, it concatenates RevGeoRQ with native Climplicit embeddings.
\end{enumerate}
Each backbone is frozen and adapted using a two-layer MLP head (64 hidden units per layer), trained for 3000 full-batch epochs with Adam and a constant learning rate of \(10^{-3}\). We use a 50/50 train-validation split across three random seeds and report the mean and standard deviation. For SatCLIP, \textit{Native} uses the best-performing \(L=40\) variant from \cite{klemmer2025satclip}, while \textit{SatCLIP-TextRQ} uses \(L=10\), which performs better in our setting; we investigate this choice in Sec. \ref{sec:ablations}.


\subsubsection{Results}

Shown in Table \ref{tab:custom-sep4}, and discussed below. 

\textbf{TextRQ vs.\ Native}: TextRQ outperforms the native frozen backbones in 16 of 28 comparisons (57\%), with the largest gains on admin-boundary aligned tasks such as country and population prediction. Results are mixed on smoother targets such as air temperature and elevation, and California Housing is a notable exception, where TextRQ improves only over GeoCLIP. As explored in Sec.~\ref{sec:ablations}, we attribute this to reverse-geocoded supervision favoring low spatial frequencies: raising SatCLIP-TextRQ to \(L=40\) recovers California Housing performance but substantially degrades every other task, so we use \(L=10\).

\textbf{Combining TextRQ and Native embeddings}: Concatenating the native and TextRQ embeddings and using this as input to the MLP yields further gains, outperforming Native in 24 of 28 comparisons (86\%). This suggests that reverse-geocoding self-supervision not only achieves competitive performance relative to continuous, naturally occurring supervision, but also provides complementary information that improves performance when the two are combined.  Similar to TextRQ, the combined models underperform on California Housing. A drawback of the combined approach is that it doubles the backbone parameter count. To reduce this overhead, we explored single-backbone multi-task pretraining with both native  and reverse-geocoded supervision; however, this consistently underperformed simple embedding concatenation.

\textbf{RevGeoRQ} (Sec. ~\ref{sec:custom-architecture}): This is the strongest single-embedding model, outperforming all Native backbones and single-embedding TextRQ models as measured by average accuracy and average $R^2$. Concatenating RevGeoRQ with Climplicit embeddings results in the best overall model. Although this combination improves aggregate performance, it reduces performance on population prediction, a task on which Climplicit itself is relatively weak. These results further supports the value of combining complementary encoder representations and supervision signals, while also highlighting that task-specific specialization may be preferable when certain downstream tasks are prioritized. Together, the two custom variants establish a new state of the art for location encoders, with one of the two achieving the best performance on 6 of 7 tasks. California Housing remains the consistent exception, with both variants underperforming on this task.

\begin{table}[t]
    \caption{Results across downstream datasets. \textbf{Bold} marks the best mean per dataset; \underline{underline} indicates a TextRQ or Combined variant that outperforms its corresponding Native model. Regression tasks report $R^2$; classification tasks report accuracy.}
    \label{tab:custom-sep4}
    \centering
    \setlength{\tabcolsep}{2pt}
    \scriptsize
    \begin{tabular}{@{}l|ccccc|cccc@{}}
    & \multicolumn{5}{c|}{\bf Regression} & \multicolumn{4}{c}{\bf Classification} \\
    \cline{2-6} \cline{7-10}
    \noalign{\vskip 3pt}
    {\bf Model} & air\_temp & elevation & population & cali\_housing & \shortstack{avg.\\[2pt]R$^2$} & country & eco regions & biome & \shortstack{avg.\\[2pt]acc.} \\
    \hline
    {\bf Baseline} & & & & & & & & & \\
    \hspace{1em}Qwen0.6B & $.820 \pm .005$ & $.727 \pm .006$ & $.692 \pm .013$ & $.675 \pm .003$ & $.729$ & $.926 \pm .003$ & $.489 \pm .007$ & $.705 \pm .009$ & $.707$ \\
    \hline
    {\bf SatCLIP} & & & & & & & & & \\
    \hspace{1em}Native & $.890 \pm .008$ & $.902 \pm .006$ & $.812 \pm .003$ & $.718 \pm .001$ & $.831$ & $.907 \pm .009$ & $.689 \pm .010$ & $.841 \pm .005$ & $.812$ \\
    \hspace{1em}TextRQ & $\underline{.906 \pm .009}$ & $\underline{.903 \pm .005}$ & $\underline{.825 \pm .010}$ & $.687 \pm .002$ & $.830$ & $\underline{.946 \pm .002}$ & $\underline{.719 \pm .013}$ & $\underline{.849 \pm .012}$ & $\underline{.838}$ \\
    \hspace{1em}Combined & $\underline{.925 \pm .008}$ & $\underline{.909 \pm .003}$ & $\underline{.821 \pm .004}$ & $.696 \pm .008$ & $\underline{.838}$ & $\underline{.944 \pm .003}$ & $\underline{.722 \pm .008}$ & $\underline{.853 \pm .009}$ & $\underline{.840}$ \\
    \hline
    {\bf GeoCLIP} & & & & & & & & & \\
    \hspace{1em}Native & $.923 \pm .003$ & $.773 \pm .005$ & $.650 \pm .013$ & $.631 \pm .016$ & $.744$ & $.907 \pm .003$ & $.575 \pm .013$ & $.799 \pm .005$ & $.760$ \\
    \hspace{1em}TextRQ & $.874 \pm .004$ & $\underline{.828 \pm .006}$ & $\underline{.782 \pm .002}$ & $\underline{.715 \pm .010}$ & $\underline{.800}$ & $.906 \pm .005$ & $.560 \pm .014$ & $.754 \pm .001$ & $.740$ \\
    \hspace{1em}Combined & $\underline{.950 \pm .004}$ & $\underline{.819 \pm .001}$ & $\underline{.719 \pm .001}$ & $\underline{.683 \pm .017}$ & $\underline{.793}$ & $\underline{.929 \pm .004}$ & $\underline{.621 \pm .010}$ & $\underline{.810 \pm .004}$ & $\underline{.787}$ \\
    \hline
    {\bf Climplicit} & & & & & & & & & \\
    \hspace{1em}Native & $.950 \pm .005$ & $.897 \pm .001$ & $.685 \pm .007$ & $\mathbf{.731 \pm .013}$ & $.816$ & $.895 \pm .004$ & $.673 \pm .011$ & $.846 \pm .004$ & $.805$ \\
    \hspace{1em}TextRQ & $.874 \pm .014$ & $.894 \pm .006$ & $\underline{.810 \pm .007}$ & $.700 \pm .005$ & $\underline{.820}$ & $\underline{.945 \pm .005}$ & $\underline{.680 \pm .010}$ & $.824 \pm .007$ & $\underline{.816}$ \\
    \hspace{1em}Combined & $\underline{.960 \pm .006}$ & $\underline{.909 \pm .004}$ & $\underline{.732 \pm .009}$ & $.722 \pm .012$ & $\underline{.831}$ & $\underline{.947 \pm .003}$ & $\underline{.711 \pm .002}$ & $\underline{.866 \pm .003}$ & $\underline{.841}$ \\
    \hline
    {\bf SINR} & & & & & & & & & \\
    \hspace{1em}Native & $.906 \pm .011$ & $.884 \pm .003$ & $.797 \pm .002$ & $.714 \pm .008$ & $.825$ & $.936 \pm .002$ & $.710 \pm .001$ & $.841 \pm .005$ & $.829$ \\
    \hspace{1em}TextRQ & $\underline{.926 \pm .007}$ & $.854 \pm .006$ & $\underline{.806 \pm .005}$ & $.670 \pm .015$ & $.814$ & $\underline{.946 \pm .002}$ & $\underline{.715 \pm .009}$ & $\underline{.850 \pm .005}$ & $\underline{.837}$ \\
    \hspace{1em}Combined & $\underline{.942 \pm .002}$ & $.882 \pm .004$ & $\underline{.811 \pm .003}$ & $.698 \pm .016$ & $\underline{.833}$ & $\underline{.945 \pm .002}$ & $\underline{.721 \pm .013}$ & $\underline{.852 \pm .007}$ & $\underline{.839}$ \\
    \hline
    {\bf Custom} & & & & & & & & & \\
    \hspace{1em}RevGeoRQ & $.929 \pm .001$ & $.896 \pm .004$ & $\mathbf{.831 \pm .004}$ & $.685 \pm .007$ & $.835$ & $\mathbf{.949 \pm .000}$ & $.723 \pm .011$ & $.852 \pm .012$ & $.841$ \\
    \hspace{1em}Combined & $\mathbf{.972 \pm .001}$ & $\mathbf{.928 \pm .012}$ & $.767 \pm .003$ & $.703 \pm .008$ & $\mathbf{.843}$ & $.945 \pm .001$ & $\mathbf{.725 \pm .004}$ & $\mathbf{.868 \pm .003}$ & $\mathbf{.846}$ \\
    \end{tabular}
\end{table}

\subsection{Location Embeddings as a geo-prior}

Prior work shows that location can provide a useful prior for geographically dependent predictions \citep{klemmer2025satclip,russwurm2024geographic}. We therefore evaluate our representations as geographic priors on iNaturalist 2018 \citep{van2018oisin}. Following prior protocols \citep{mac2019presence,russwurm2024geographic,dhakal2025range}, for each image \(I_i\), a pretrained image classifier produces \(P(y\mid I_i)\), while a linear classifier trained on location embedding \(c_i\) produces \(P(y\mid c_i)\). At inference, we combine them multiplicatively as \(P(y\mid I_i,c_i)\propto P(y\mid I_i)P(y\mid c_i)\) and report top-\(k\) accuracy. We exclude SINR because its location encoder is supervised with iNaturalist species labels, preventing an independent transfer evaluation. Additional details are provided in Appendix~\ref{app:inat}.

Tab.~\ref{tab:inat-sep25} show that TextRQ generally improves over Native, outperforming all  backbones at top-1, matching or exceeding them at top-3, and improving top-5 for SatCLIP and GeoCLIP; only Climplicit retains a Native advantage at top-5. Combining Native and TextRQ helps only the weaker GeoCLIP backbone. RevGeoRQ outperforms the vision-only baseline and native GeoCLIP but trails the strongest adapted TextRQ models. SatCLIP-TextRQ is the best overall based on mean top-$k$.

\begin{table}[t]
    \caption{iNaturalist 2018 geo-prior evaluation. Values report top-$k$ validation accuracy as mean $\pm$ std over three seeds; the \textbf{mean} column gives the per-row average over top-1, top-3, and top-5. \textbf{Bold} marks the best mean within each model family; \underline{underline} indicates improvement over the corresponding Native baseline.}
    \label{tab:inat-sep25}
    \centering
    \setlength{\tabcolsep}{6pt}
    \renewcommand{\arraystretch}{0.9}
    \scriptsize
    \begin{tabular}{l cccc}
    {\bf Model} & top-1 & top-3 & top-5 & mean \\
    \hline
    {\bf Baseline} & & & & \\
    \hspace{1em}Vision-only (Inception-v3) & $0.6617$ & $0.8331$ & $0.8803$ & $0.7917$ \\
    \hline
    {\bf SatCLIP} & & & & \\
    \hspace{1em}Native (L=40) & $0.7537 \pm 0.0009$ & $0.8945 \pm 0.0003$ & $0.9280 \pm 0.0003$ & $0.8587$ \\
    \hspace{1em}TextRQ (L=10) & $\underline{\mathbf{0.7730 \pm 0.0008}}$ & $\underline{\mathbf{0.9019 \pm 0.0006}}$ & $\underline{\mathbf{0.9306 \pm 0.0004}}$ & $\underline{\mathbf{0.8685}}$ \\
    \hspace{1em}Combined      & $\underline{0.7677 \pm 0.0007}$ & $\underline{0.8964 \pm 0.0006}$ & $0.9256 \pm 0.0003$ & $\underline{0.8633}$ \\
    \hline
    {\bf GeoCLIP} & & & & \\
    \hspace{1em}Native   & $0.6901 \pm 0.0001$ & $0.8546 \pm 0.0001$ & $0.8975 \pm 0.0000$ & $0.8141$ \\
    \hspace{1em}TextRQ   & $\underline{0.7584 \pm 0.0003}$ & $\underline{0.8985 \pm 0.0005}$ & $\underline{0.9291 \pm 0.0008}$ & $\underline{0.8620}$ \\
    \hspace{1em}Combined & $\underline{\mathbf{0.7647 \pm 0.0003}}$ & $\underline{\mathbf{0.9025 \pm 0.0005}}$ & $\underline{\mathbf{0.9320 \pm 0.0008}}$ & $\underline{\mathbf{0.8664}}$ \\
    \hline
    {\bf Climplicit} & & & & \\
    \hspace{1em}Native   & $0.7653 \pm 0.0005$ & $0.8938 \pm 0.0004$ & $\mathbf{0.9237 \pm 0.0001}$ & $0.8609$ \\
    \hspace{1em}TextRQ   & $\underline{\mathbf{0.7669 \pm 0.0010}}$ & $\underline{\mathbf{0.8938 \pm 0.0006}}$ & $0.9222 \pm 0.0002$ & $\underline{\mathbf{0.8610}}$ \\
    \hspace{1em}Combined & $0.7604 \pm 0.0005$ & $0.8842 \pm 0.0005$ & $0.9138 \pm 0.0005$ & $0.8528$ \\
    \hline
    {\bf Custom} & & & & \\
    \hspace{1em}RevGeoRQ & $\mathbf{0.7545 \pm 0.0002}$ & $\mathbf{0.8808 \pm 0.0007}$ & $\mathbf{0.9104 \pm 0.0004}$ & $\mathbf{0.8486}$ \\
    \hspace{1em}Combined & $0.7489 \pm 0.0011$ & $0.8706 \pm 0.0002$ & $0.8998 \pm 0.0007$ & $0.8398$ \\
    \end{tabular}
\end{table}

\section{Ablation Study}\label{sec:ablations}

Table~\ref{tab:ablation-main} ablates key architectural choices. Variants with a location encoder use the adapted SatCLIP backbone, except RevGeoRQ variants, which follow Sec.~\ref{sec:downstream-finetuning}.

\textbf{Text encoder and RQ-VAE reconstruction loss}: Qwen-0.6B alone performs best on tasks aligned with administrative names, especially country classification, but is weaker on finer-grained spatial tasks and poorest on ecoregions, whose 649 classes often cross administrative boundaries. Adding RQ-VAE layers with DIRECT latitude--longitude encoding \citep{russwurm2024geographic} improves air temperature, ecoregion, and biome performance, but hurts elevation and California housing. Adding the SatCLIP backbone yields the full TextRQ model (Sec.~\ref{sec:downstream-finetuning}), combining reverse-geocoded semantics with continuous geographic features and substantially improving overall performance. Replacing Qwen-0.6B with Qwen-4B provides further but modest gains.

\textbf{Effect of spatial frequency}: Increasing the spherical-harmonic degree from \(L=10\) to \(L=40\) degrades all tasks except California Housing, where \(L=40\) performs best. Consistent with \cite{klemmer2025satclip}, higher \(L\) supports finer spatial variation, benefiting tasks such as housing prices that change over small areas, while smoother targets such as temperature and elevation gain little. We hypothesize that reverse-geocoded supervision further favors lower frequencies: it assigns shared representations to large, sometimes heterogeneous regions, making \(L=10\) a better match to both the training signal and predominantly low-frequency downstream tasks.

\textbf{Effect of concatenation}: Concatenating TextRQ embeddings with native location encoder embeddings generally improves performance, suggesting that our administrative-boundary-based text supervision provides a complementary signal to the naturally occurring supervision used by the native backbones. Moreover, the relative strength of the native backbones is largely preserved after concatenation: when one native embedding outperforms another on a given task, its corresponding concatenated representation typically remains stronger as well. 

\begin{table}[t]
    \caption{Model architecture and parameter ablations. Regression reports \(R^2\), classification reports accuracy, and \textbf{avg.} denotes the row mean within each group. \textbf{Bold} marks the best value per column.}
    \label{tab:ablation-main}
    \centering
    \setlength{\tabcolsep}{1.25pt}
    \scriptsize
    \begin{tabular}{@{}l|ccccc|cccc@{}}
    & \multicolumn{5}{c|}{\bf Regression} & \multicolumn{4}{c}{\bf Classification} \\
    \cline{2-6} \cline{7-10}
    \noalign{\vskip 3pt}
    {\bf Model} & air\_temp & elevation & population & cali\_housing & \shortstack{avg.\\[2pt]R$^2$} & country & eco regions & biome & \shortstack{avg.\\[2pt]acc.} \\
    \hline
    Qwen-0.6B & $.820 \pm .005$ & $.727 \pm .006$ & $.692 \pm .013$ & $.675 \pm .003$ & $.729$ & $.926 \pm .003$ & $.489 \pm .007$ & $.705 \pm .009$ & $.707$ \\
    \begin{tabular}[c]{@{}l@{}}RQVAE-only\\[-2pt](0.6B)\end{tabular}
    & $.913 \pm .005$ & $.564 \pm .165$ & $.683 \pm .049$ & $.508 \pm .021$ & $.667$ & $.907 \pm .010$ & $.560 \pm .022$ & $.771 \pm .005$ & $.746$ \\
    \begin{tabular}[c]{@{}l@{}}SatCLIP-TextRQ\\[-2pt](L=40, 0.6B)\end{tabular}
    & $-.028 \pm .019$ & $.649 \pm .019$ & $.458 \pm .038$ & $\mathbf{.729 \pm .013}$ & $.452$ & $.382 \pm .012$ & $.266 \pm .003$ & $.415 \pm .008$ & $.354$ \\
    \begin{tabular}[c]{@{}l@{}}SatCLIP-TextRQ\\[-2pt](L=10, 0.6B)\end{tabular}
    & $.906 \pm .009$ & $.903 \pm .005$ & $.825 \pm .010$ & $.687 \pm .002$ & $.830$ & $.946 \pm .002$ & $.719 \pm .013$ & $.849 \pm .012$ & $.838$ \\
    \begin{tabular}[c]{@{}l@{}}SatCLIP-TextRQ\\[-2pt](L=10, 4B)\end{tabular}
    & $.922 \pm .005$ & $.890 \pm .011$ & $.831 \pm .001$ & $.682 \pm .019$ & $.831$ & $.949 \pm .002$ & $.725 \pm .005$ & $.853 \pm .009$ & $.842$ \\
    RevGeoRQ & $.929 \pm .001$ & $.896 \pm .004$ & $.831 \pm .004$ & $.685 \pm .007$ & $.835$ & $\mathbf{.949 \pm .000}$ & $.723 \pm .011$ & $.852 \pm .012$ & $.841$ \\
    \begin{tabular}[c]{@{}l@{}}RevGeoRQ +\\[-2pt]SatCLIP-L40\end{tabular}
    & $.940 \pm .003$ & $.907 \pm .004$ & $\mathbf{.834 \pm .005}$ & $.673 \pm .010$ & $.839$ & $.947 \pm .004$ & $\mathbf{.727 \pm .009}$ & $.856 \pm .006$ & $.843$ \\
    \begin{tabular}[c]{@{}l@{}}RevGeoRQ +\\[-2pt]Climplicit\end{tabular}
    & $\mathbf{.972 \pm .001}$ & $\mathbf{.928 \pm .012}$ & $.767 \pm .003$ & $.703 \pm .008$ & $\mathbf{.843}$ & $.945 \pm .001$ & $.725 \pm .004$ & $\mathbf{.868 \pm .003}$ & $\mathbf{.846}$ \\
    \end{tabular}
\end{table}

\textbf{Objective function}: Using a CLIP-style contrastive loss to align the geospatial embedding with the Qwen text embedding shows equivalent or better performance to the RQVAE loss for SatCLIP and SINR, but significantly worse performance for GeoCLIP and Climplicit. Due to its better cross-encoder generalizability, we choose the RQVAE loss/architecture. Details in Appendix \ref{app:ablations} Table~\ref{tab:ablating-objective}.

We also tuned codebook size, number of levels, and latent dimension (final settings in Appendix~\ref{app:rqvae}).

\section{Embeddings analysis}
\label{sec:embedding-analysis}

\subsection{Semantic IDs}

\begin{figure}[h]
    \begin{center}
        \includegraphics[width=1.0\linewidth]{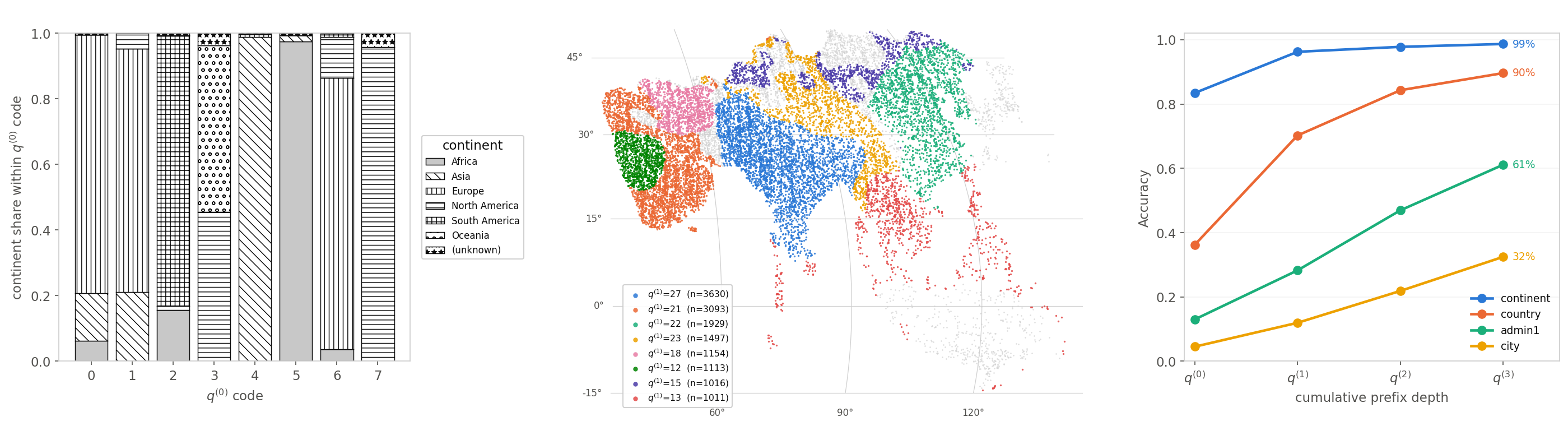}    
    \end{center}
    \caption{\textbf{Left:} the reverse-geocoded continent share within each $q^{(0)}$ code. \textbf{Center:} the eight largest $q^{(1)}$ codes within $q^{(0)}=4$. \textbf{Right:} attribute accuracy vs.\ cumulative prefix depth.}
    \label{fig:l1-map-and-q0-dist-and-attr-acc}
\end{figure}

\textbf{Emergent geographic hierarchy}: Similar to prior work \citep{rajput2023recommender, li2026tokens}, we observe that the hierarchical nature of RQ-VAEs induces a coarse-to-fine structure in the learned Semantic IDs; in our case, this coarse-to-fine structure manifests as a geographic multiresolution zoom. To illustrate this, we train a SatCLIP TextRQ model as in Sec.~\ref{sec:pretrain-adapting-backbones}, but with codebook sizes 8, 32, 128, and 256 for 1000 epochs, and use this model to compute the Semantic IDs for the S2-100K points. As shown in Fig.~\ref{fig:arch-and-q0} (Right), the coarsest codebook $\mathcal{C}_{0}$ with selected codes $q^{(0)}$ partitions the latitude–longitude space into eight regions at continental scale. Cross-tabulating each $\mathcal{C}_{0}$ code against its reverse-geocoded continent labels shows that every code roughly aligns with a single continent or a coherent macro-region (Fig.~\ref{fig:l1-map-and-q0-dist-and-attr-acc} Left), showing that the coarsest level estimates continent-scale semantics. The next codebook $\mathcal{C}_{1}$ refines each coarse region into finer, still-contiguous sub-regions. Fig.~\ref{fig:l1-map-and-q0-dist-and-attr-acc}~(Center) shows this within the South and Central Asian region ($\mathcal{C}_{0}{=}4$), where the eight largest $\mathcal{C}_{1}$ codes carve out coherent sub-continental clusters---South Asia, the Arabian Peninsula, East Asia, and Southeast Asia---each dominated by a single country or country group. The codes thus form a nested spatial hierarchy that zooms from continents ($\mathcal{C}_{0}$) to sub-continental regions ($\mathcal{C}_{1}$) learned implicitly through the Qwen-encoded reverse geocoded strings.

\textbf{Admin Boundary Aware Semantic IDs}: We measure how well a Semantic ID predicts a point's administrative attributes at an unseen location. From the training split, we record the majority attribute label for every cumulative code prefix — $q^{(0)}$, then $q^{(0)}q^{(1)}$, up to $q^{(0)}q^{(1)}q^{(2)}q^{(3)}$. Then, for each point in the validation split, we form its Semantic ID and predict the majority label recorded in training for its prefix at that depth to compare with the ground truth attribute. If the prefix never occurs in training, we back off to the next-shorter prefix — recursively, down to $q^{(0)}$ and finally the global majority. We report accuracy at each prefix depth, which quantifies how much locational information the codes carry. Accuracy rises monotonically with prefix depth for every attribute (Fig.~\ref{fig:l1-map-and-q0-dist-and-attr-acc} Right), showing that the codes carry encode coarse-to-fine boundary awareness: continent is largely fixed by $q^{(0)}$ alone (83\%) and country by $q^{(0)}q^{(1)}$ (70\%), while admin1 and city keep improving up to the full Semantic ID (61\% and 33\%).  Admin1 and city lag; this is not a sign that the Semantic ID lacks city/admin1 information, but rather the fact that even the finest cell (368 km - explored in Appendix~\ref{app:semid-additional}) is larger than most cities. Therefore, one cell covers many cities and a majority vote can't separate them; this may be mitigated by using larger codebooks at deeper levels.

We additionally study the spatial resolution induced by the codebook prefixes, whose cell spread shrinks from 4244 to 368\,km over successive levels; details are provided in Appendix~\ref{app:semid-additional}.

\subsection{Location Encoder Outputs}

\begin{figure}[h]
    \begin{center}
        \includegraphics[width=1.0\linewidth]{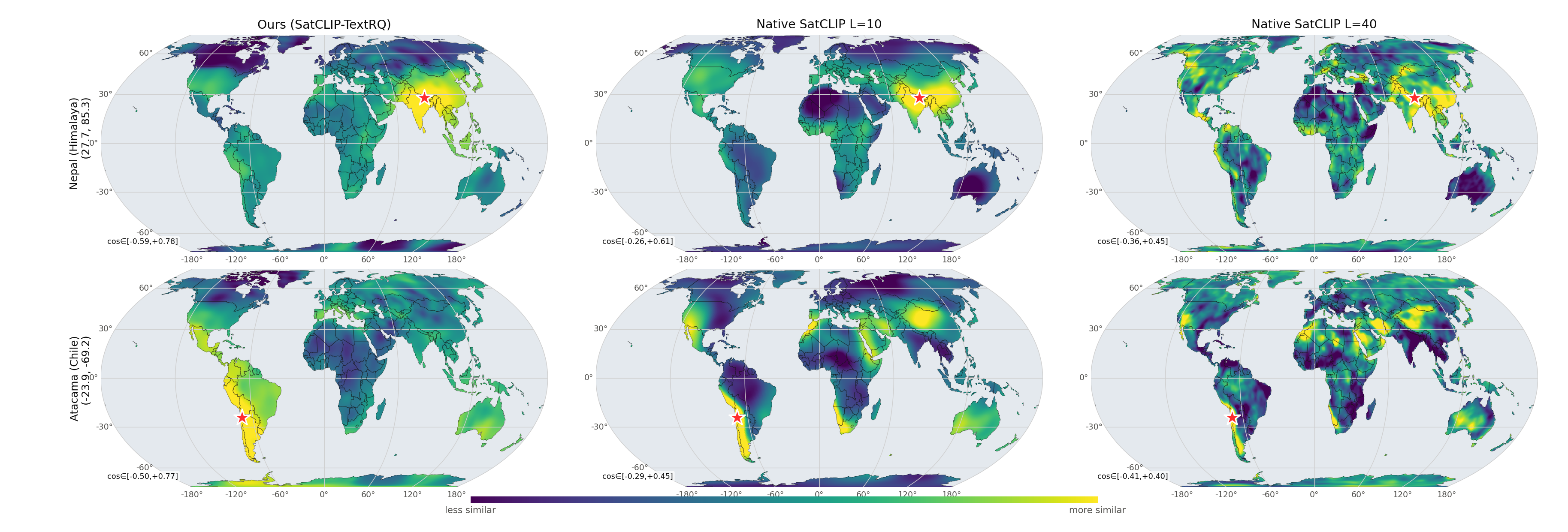}    
    \end{center}
    \caption{Cosine similarity of each land embedding to a query point (\textcolor{red}{$\star$}) --- a Himalayan query (\textbf{top}) and an Atacama query (\textbf{bottom}) --- for our \textbf{TextRQ} encoder (left) vs.\ native \textbf{SatCLIP} L=10 and L=40 (center/right).}
        \label{fig:loc-emb-viz}
\end{figure}

Since the location-encoder outputs feed the downstream tasks, we qualitatively examine what this layer has learned. As illustration, we mark a query point in the Himalayan range (Nepal, 27.7°N 85.3°E) and one in the Atacama desert (23.9°S 69.2°W), and color every land location on a dense global 0.5° grid by its cosine similarity to that query point. These two cases are shown at the top and bottom of Fig. \ref{fig:loc-emb-viz}, for our TextRQ encoder alongside the native SatCLIP L=10 and L=40 encoders.

All three encoders light up the immediate neighborhood of every query, confirming that location encoders learn the basic prior that spatially nearby points are similar (spatial autocorrelation), regardless of training signal. Beyond that, the fields diverge. Our TextRQ model's similar regions align cleanly with administrative structure — bright at the country level, tapering to medium similarity across the surrounding subcontinent and then the continent — evidencing a well-formed country → subcontinent → continent hierarchy; the Himalaya query lights Nepal and South Asia coherently, and the Atacama query lights Chile and the Andean/southern-cone belt. SatCLIP does not show this: points within the same country can read as dissimilar, and this fragmentation intensifies with spherical-harmonic degree (worse in the high-frequency L=40 than in L=10). In exchange, SatCLIP, which is trained on imagery, recovers long-range, cross-continental similarities from shared physical character: the Himalaya query also lights the Andes and Rockies, and the Atacama query lights other arid west-coast deserts worldwide; these are largely missed by our TextRQ embedding. In summary, TextRQ captures administrative hierarchies while image-trained encoders like SatCLIP capture natural, physical characteristics; both learn spatial autocorrelation well. These are complementary signals, and combining them improves over either alone, as shown in Table \ref{tab:custom-sep4}.

\section{Limitations}
\label{sec:limitations}

A limitation of reverse-geocoded supervision is that distinct locations within the same administrative boundaries may map to identical strings, even when they are far apart or differ substantially in local characteristics; we analyze this duplication effect in Appendix~\ref{app:spatial-granularity}. The hierarchical discretization of our model can also obscure non-local similarities, particularly between geographically distant regions on different continents. Finally, performance depends on reverse-geocoding quality. More precise geocoders may recover finer distinctions, but coverage remains inherently limited in large, sparsely populated regions where neighborhood- and locality-level boundaries are coarse or absent.

\section{Conclusion}
\label{sec:conclusions}

We introduce reverse-geocoded text as a cheap, scalable self-supervision signal that distills
administrative structure into compact location encoders without inference-time LLM or geocoding cost.
Our method improves existing backbones and also achieves gains when used in custom optimized RevGeoRQ architecture, showing that administrative supervision complements
naturally-occurring and continuous signals. RevGeoRQ is the strongest single-embedding model in our experiments; combining it with Climplicit achieves the best overall results. The learned Semantic IDs form an interpretable hierarchy, continent-scale down to local regions. Future work includes using finer-grained geocoders, experimenting with larger codebooks, and extending Semantic IDs beyond
pure location encoding to search, retrieval, and generative applications.







\bibliography{iclr2027_conference}
\bibliographystyle{iclr2027_conference}

\appendix

\section{Residual-quantized variational autoencoder design}
\label{app:rqvae}

\paragraph{Encoder/decoder dimensions.}
The encoder $\mathcal{E}_q$ and decoder $\mathcal{D}_q$ are symmetric MLPs,
$\mathcal{E}_q:1024\!\rightarrow\!512\!\rightarrow\!256\!\rightarrow\!128\!\rightarrow\!32$
and
$\mathcal{D}_q:32\!\rightarrow\!128\!\rightarrow\!256\!\rightarrow\!512\!\rightarrow\!1024$.

\paragraph{Codebook initialization.}
To prevent codebook collapse, where most inputs are mapped to only a few
codewords, each level's codebook is initialized with k-means centroids computed on
the residuals of the first training batch \citep{zeghidour2021soundstream}.

\paragraph{Hyperparameters.}
We set $d_t=1024$, $d_q=32$, $m=4$ levels, $K=32$ codewords per level, and
$\beta=0.25$; the resulting four-level, 32-entry codebooks index up to
$K^{m}=32^{4}\approx1.05\times10^{6}$ distinct Semantic IDs.

\section{Pretraining Details}
    \label{app:pretraining-details}

    \begin{table}[H]
        \caption{Per-encoder pretraining setup}
        \label{tab:pretrain-params}
        \begin{center}
        \setlength{\tabcolsep}{5pt}
        \footnotesize
        \begin{tabular}{l ccc p{2.6cm}}
        & {\bf SatCLIP} & {\bf GeoCLIP} & {\bf Climplicit} & {\bf SINR} \\
        \hline \\
        Backbone variant & SatCLIP ViT16-L10 & GeoCLIP & Climplicit & $\mathcal{L}_{\text{AN-full}}$ SINR (Coords., 1000/class) \\
        Training data     & S2-100K           & MP-16 Flickr       & CHELSA land grid & iNaturalist \\
        \hspace{1em}(\# points) & (91{,}858) & ($\sim$4.72M) & ($\sim$224M) & ($\sim$35.5M) \\
        Budget & \texttt{max\_epochs}=100 & \texttt{max\_steps}=50k & \texttt{max\_steps}=50k & \texttt{max\_epochs}=50 \\[3pt]
        \hline \\
        Optimizer & \multicolumn{4}{c}{AdamW (betas $0.9,0.999$; eps $10^{-8}$)} \\
        LR & \multicolumn{4}{c}{peak $3\times10^{-4}$, weight decay $10^{-4}$} \\
        Scheduler & \multicolumn{4}{c}{linear warmup (5\% of steps) $\to$ cosine decay to 10\% floor} \\
        Batch size & \multicolumn{4}{c}{4096} \\
        \end{tabular}
        \end{center}
    \end{table}

\section{Contrastive vs.\ RQ-VAE Objectives Across Backbones}
    \label{app:ablations}

    Shown in Table~\ref{tab:ablating-objective}.
    
\begin{table}[H]
    \caption{Downstream performance after ablating the pretraining objective on each encoder's own data. \textbf{Bold} shows best per backbone-task pair between CLIP and RQVAE objectives}
    \label{tab:ablating-objective}
    \centering
    \setlength{\tabcolsep}{1.5pt}
    \scriptsize
    \begin{tabular}{@{}l|ccccc|cccc@{}}
    & \multicolumn{5}{c|}{\bf Regression} & \multicolumn{4}{c}{\bf Classification} \\
    \cline{2-6} \cline{7-10}
    \noalign{\vskip 3pt}
    \shortstack[l]{\textbf{Model}\\[2pt]\textbf{(all TextRQ)}} & air\_temp & elevation & population & \shortstack{cali\\[2pt]housing} & \shortstack{avg.\\[2pt]R$^2$} & country & \shortstack{eco\\[2pt]regions} & biome & \shortstack{avg.\\[2pt]acc.} \\
    \hline
    {\bf SatCLIP (L=10)} & & & & & & & & & \\
    \hspace{1em}CLIP  & $\mathbf{.930 \pm .009}$ & $.888 \pm .007$ & $\mathbf{.831 \pm .002}$ & $.671 \pm .006$ & $\mathbf{.830}$ & $\mathbf{.948 \pm .004}$ & $\mathbf{.723 \pm .009}$ & $\mathbf{.859 \pm .007}$ & $\mathbf{.843}$ \\
    \hspace{1em}RQVAE & $.906 \pm .009$ & $\mathbf{.903 \pm .005}$ & $.825 \pm .010$ & $\mathbf{.687 \pm .002}$ & $\mathbf{.830}$ & $.946 \pm .002$ & $.719 \pm .013$ & $.849 \pm .012$ & $.838$ \\
    \hline
    {\bf GeoCLIP} & & & & & & & & & \\
    \hspace{1em}CLIP  & $.754 \pm .010$ & $.436 \pm .044$ & $.253 \pm .027$ & $.694 \pm .007$ & $.534$ & $.767 \pm .003$ & $.414 \pm .001$ & $.651 \pm .004$ & $.611$ \\
    \hspace{1em}RQVAE & $\mathbf{.796 \pm .022}$ & $\mathbf{.693 \pm .007}$ & $\mathbf{.652 \pm .003}$ & $\mathbf{.702 \pm .003}$ & $\mathbf{.711}$ & $\mathbf{.793 \pm .006}$ & $\mathbf{.417 \pm .008}$ & $\mathbf{.666 \pm .006}$ & $\mathbf{.625}$ \\
    \hline
    {\bf ReSIREN} & & & & & & & & & \\
    \hspace{1em}CLIP  & $.573 \pm .044$ & $\mathbf{.896 \pm .004}$ & $.767 \pm .003$ & $\mathbf{.705 \pm .008}$ & $.735$ & $.913 \pm .006$ & $.652 \pm .005$ & $.808 \pm .008$ & $.791$ \\
    \hspace{1em}RQVAE & $\mathbf{.874 \pm .014}$ & $.894 \pm .006$ & $\mathbf{.810 \pm .007}$ & $.700 \pm .005$ & $\mathbf{.820}$ & $\mathbf{.945 \pm .005}$ & $\mathbf{.680 \pm .010}$ & $\mathbf{.824 \pm .007}$ & $\mathbf{.816}$ \\
    \hline
    {\bf SINR} & & & & & & & & & \\
    \hspace{1em}CLIP  & $.905 \pm .014$ & $\mathbf{.867 \pm .005}$ & $\mathbf{.812 \pm .007}$ & $\mathbf{.758 \pm .001}$ & $\mathbf{.836}$ & $.942 \pm .003$ & $.703 \pm .008$ & $.842 \pm .007$ & $.829$ \\
    \hspace{1em}RQVAE & $\mathbf{.926 \pm .007}$ & $.854 \pm .006$ & $.806 \pm .005$ & $.670 \pm .015$ & $.814$ & $\mathbf{.946 \pm .002}$ & $\mathbf{.715 \pm .009}$ & $\mathbf{.850 \pm .005}$ & $\mathbf{.837}$ \\
    \end{tabular}
\end{table}

\section{iNaturalist Geo-prior Finetuning}
    \label{app:inat}
    For the iNaturalist 2018 geo-prior evaluation, we train the location probe for 30 epochs using Adam with an initial learning rate of $5\times10^{-4}$, \texttt{ExponentialLR} with $\gamma=0.98$, batch size 1024, \texttt{pos\_weight}=2048, and a background bank of 100k locations. Following prior work \citep{russwurm2024geographic,dhakal2025range}, we use the full ``assume negative'' loss of \citep{cole2023spatial}, which is designed for the presence-only annotations in iNaturalist. Results are reported as mean $\pm$ standard deviation over seeds $\{0,1,2\}$; the seed controls probe initialization, background-bank sampling, and minibatch ordering, while cached frozen embeddings are shared across seeds.

\section{Spatial Resolution of Semantic IDs}
    \label{app:semid-additional}

\begin{figure}[h]
    \centering
    \includegraphics[width=0.65\linewidth]{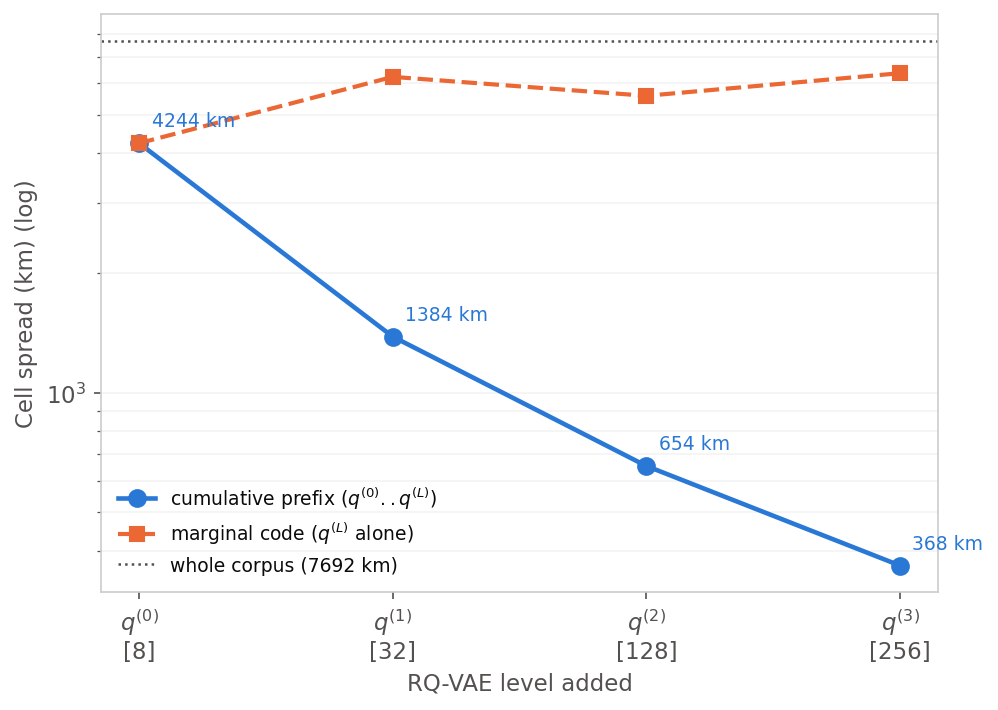}
    \caption{Cell spread per level (cumulative vs.\ marginal).}
    \label{fig:cell-spread}
\end{figure}

We summarize the spatial resolution induced by each Semantic ID prefix using the geographic dispersion of points assigned to the same prefix. For each cumulative prefix depth (\(q^{(0)}\), \(q^{(0)}q^{(1)}\), ..., \(q^{(0)}q^{(1)}q^{(2)}q^{(3)}\)), we group points sharing that prefix into a cell, map each point to a unit vector on the sphere, and take the mean vector as the cell's spherical centroid. We then measure the great-circle (haversine) distance from each point to its cell centroid ($R=6371\,\mathrm{km}$) and report the point-weighted root-mean-square of these distances as the cell dispersion or spread at that depth. As baselines, we apply the same calculation to marginal codes \(\mathcal{C}_1,\mathcal{C}_2,\mathcal{C}_3\) independently and to the entire dataset treated as a single cell. The cumulative cell dispersion shrinks monotonically from $4244$ to $1384$ to $654$ to $368\,\mathrm{km}$ ($\sim\!2$--$3\times$ per level), i.e.\ the Semantic ID acts as a coarse-to-fine geographic zoom from continent to province scale (Fig.~\ref{fig:cell-spread}). Individual marginal codes instead stay pinned near the whole-corpus spread ($7692\,\mathrm{km}$), confirming that a single level carries no standalone location and that only the nested prefix denotes a place. The $368\,\mathrm{km}$ floor at full depth coincides with the city-probe ceiling, since a city is far smaller than the finest cell.

\section{Loss of Spatial Granularity from Reverse Geocoding}
    \label{app:spatial-granularity}

Self-supervised labels can be no more precise than the observation that produces them. The spatial resolution of our text targets is therefore bounded by the reverse geocoder, not by the input coordinates, and our approach depends heavily on the geocoder's quality. We use the free, offline \texttt{reverse\_geocoder} Python package, which snaps each coordinate to its nearest GeoNames populated place and offers only moderate coverage and precision. Commercial geocoders with finer resolution and broader coverage are costly at the scale of hundreds of millions of points. As Table~\ref{tab:geocode-reduction} shows, this maps between 92K and 224M distinct input locations onto at most ${\sim}$101K text targets, a reduction of 4.3$\times$ (S2-100K) to 2{,}375$\times$ (CHELSA), so all points near the same place receive an identical supervisory signal. The cost appears in Table~\ref{tab:custom-sep4} on tasks that depend on fine-grained local structure or smooth spatial variation. On California Housing, TextRQ falls below its Native counterpart for SatCLIP (.718$\rightarrow$.687), ReSIREN (.731$\rightarrow$.700), and SINR (.714$\rightarrow$.670), and none of these Combined variants recovers the Native score. For air temperature, TextRQ drops sharply for ReSIREN (.950$\rightarrow$.874), whose training set has the largest reduction rate, and for GeoCLIP (.923$\rightarrow$.874), while elevation declines for SINR (.884$\rightarrow$.854). By contrast, tasks aligned with administrative units benefit: population improves for every backbone and country for three of four. TextRQ thus inherits its geocoder's granularity: it performs well where the signal changes between places, especially at the step-function transitions typically induced by administrative boundaries, but loses information where the signal varies within a place.

\begin{table}[H]
    \caption{Input-space reduction from reverse geocoding. The reduction rate is the number of distinct points divided by the number of distinct reverse geocoded strings, i.e.\ how many distinct input locations collapse onto each text target.}
    \label{tab:geocode-reduction}
    \centering
    \setlength{\tabcolsep}{2pt}
    \scriptsize
    \begin{tabular}{@{}l|rrrr@{}}
    {\bf Model (dataset)} & Total points & Distinct points & \shortstack{Distinct reverse\\[2pt]geocoded strings} & \shortstack{Reduction rate\\[2pt](distinct points per string)} \\
    \hline
    SatCLIP (S2-100K) & 91,858 & 91,858 & 21,188 & 4.3$\times$ \\
    GeoCLIP (MP-16) & 4,723,695 & 2,875,487 & 81,997 & 35.1$\times$ \\
    Climplicit (CHELSA) & 224,368,589 & 224,368,589 & 94,473 & 2,375.0$\times$ \\
    SINR (iNat) & 35,500,262 & 27,862,308 & 101,411 & 274.7$\times$ \\
    \end{tabular}
\end{table}

\end{document}

%% file: math_commands.tex
\usepackage{amsmath,amsfonts,bm}

\def\eqref#1{equation~\ref{#1}}

\def\1{\bm{1}}

\DeclareMathAlphabet{\mathsfit}{\encodingdefault}{\sfdefault}{m}{sl}
\SetMathAlphabet{\mathsfit}{bold}{\encodingdefault}{\sfdefault}{bx}{n}

